\documentclass[11pt]{article}
\usepackage{acl2015}
\usepackage{times}
\usepackage{url}
\usepackage{latexsym}
\usepackage{lscape}
\usepackage{color}
\usepackage{soul}

\usepackage{epstopdf}
\usepackage{graphicx} 

\title{Improving distant supervision using inference learning}

\author{Roland Roller$^{1}$,~~~Eneko Agirre$^{2}$,~~~Aitor Soroa$^{2}$ \and Mark Stevenson$^{1}$\\
  $^{1}$ Department of Computer Science, University of Sheffield\\
   {\tt roland.roller,mark.stevenson@sheffield.ac.uk}\\
  $^{2}$ IXA NLP group, University of the Basque Country\\
 {\tt e.agirre,a.soroa@ehu.eus}}

\date{}

\begin{document}
\maketitle
\begin{abstract}

Distant supervision is a widely applied approach to automatic training of relation extraction systems and has the advantage that it can generate large amounts of labelled data with minimal effort. However, this data may contain errors and consequently systems trained using distant supervision tend not to perform as well as those based on manually labelled data. This work proposes a novel method for detecting potential false negative training examples using a knowledge inference method. Results show that our approach improves the performance of relation extraction systems trained using distantly supervised data. 
\end{abstract}

\section{Introduction}

Distantly supervised relation extraction relies on automatically labelled data generated using information from a knowledge base. A sentence is annotated as a positive example if it contains a pair of entities that are related in the knowledge base. Negative training data is often generated using a closed world assumption: pairs of entities not listed in the knowledge base are assumed to be unrelated and sentences containing them considered to be negative training examples. However this assumption is violated when the knowledge base is incomplete which can lead to sentences containing instances of relations being wrongly annotated as negative examples.


We propose a method to improve the quality of distantly supervised data by identifying  possible wrongly annotated negative instances. Our proposed method includes a version of the Path Ranking Algorithm (PRA) \cite{Lao:2010,Lao:2011} which infers relation paths by combining random walks though a knowledge base. We use this knowledge inference to detect possible false negatives (or at least  entity pairs closely connected to a target relation) in automatically labelled training data and show that their removal can improve relation extraction performance.


\section{Related Work}


Distant supervision is widely used to train relation extraction systems with Freebase and Wikipedia commonly being used as knowledge bases, e.g. \cite{Mintz:2009,Riedel:2010,Krause:2012,Zhang:2013,Min13,Rit13}.
The main advantage is its ability to automatically generate large amounts of training data automatically. On the other hand, this automatically labelled data is noisy and usually generates lower performance than approaches trained using manually labelled data. 
A range of filtering approaches have been applied to address this problem including multi-class SVM \cite{Nguyen:2011a} and Multi-Instance learning methods \cite{Riedel:2010,Surdeanu:2012}. These approaches take into account the fact that entities might occur in different relations at the same time and may not necessarily express the target relation. Other approaches focus directly on the noise in the data. For instance \newcite{Takamatsu:2012} use a generative model to predict incorrect data while \newcite{Intxaurrondo:2013} use a range of heuristics including PMI to remove noise. \newcite{Augenstein:2014b} apply techniques to detect highly ambiguous entity pairs and discard them from their labelled training set.

This work proposes a novel approach to the problem by applying an inference learning method to identify potential false negatives in distantly labelled data. 
Our method makes use of a modified version of PRA to learn 
relation paths from a knowledge base and uses this information to identify false negatives.

\section{Data and Methods}

We chose to apply our approach to relation extraction tasks from the biomedical domain since this has proved to be an important problem within these documents \cite{Jen06,Hah12,Coh13,Roller:2014b}. In addition, the first application of distant supervision was to biomedical journal articles \cite{Craven:1999}. In addition, the most widely used knowledge source in this domain, the UMLS Metathesaurus \cite{Bod04}, is an ideal resource to apply inference learning given its rich structure. 

We develop classifiers to identify relations found in two subsets of UMLS: the National Drug File-Reference Terminology (ND-FRT) and the National Cancer Institute Thesaurus (NCI). A corpus of approximately 1,000,000 publications is used to create the distantly supervised training data. The corpus contains abstracts published between 1990 and 2001 annotated with UMLS concepts using MetaMap \cite{Aronson:2010}.

\subsection{Distantly labelled data}\label{sec:distantly_labelled_data}




Distant supervision is carried out for a target UMLS relation by identifying instance pairs and using them to create a set of positive instance pairs. Any pairs which also occur as an instance pair of another UMLS relation are removed from this set. A set of negative instance pairs is then created by forming new combinations that do not occur within the positive instance pairs. Sentences containing a positive or negative instance pair are then extracted to generate positive and negative training examples for the relation. These candidate sentences are then stemmed \cite{Porter:1997} and PoS tagged \cite{Charniak:2005}. 

The sets of positive and negative training examples are then filtered to remove sentences that meet any of the following criteria: contain the same positive pair more than once; contain both a positive and negative pair; more than 5 words between the two elements of the instance pair; contain very common instance pairs.


\subsection{PRA-Reduction}\label{sec:pra_reduction}

PRA \cite{Lao:2010,Lao:2011} is an algorithm that infers new relation instances from knowledge bases. By considering a knowledge base as a graph, where nodes are connected through typed relations, it performs random walks over it and finds bounded-length relation paths that connect graph nodes. These paths are used as features in a logistic regression model, which is meant to predict new relations in the graph. Although initially conceived as an algorithm to discover new links in the knowledge base, PRA can also be used to learn relevant relation paths for any given relation. For instance, if \textit{x} and \textit{y} are related via \textit{sibling} relation, the model trained by PRA would learn that the relation path \textit{parent(x,a) $\wedge$ \_parent(a,y)}\footnote{An underline (`\_') prefix represents the inverse of a relation while $\wedge$ represents path composition.} is highly relevant, as siblings share the same parents.

%

Knowledge graphs were extracted from the ND-FRT and NCI vocabularies generating approximately $200,000$ related instance pairs for ND-FRT and $400,000$ for NCI. PRA is then run on both graphs in order to learn paths for each target relation. Table \ref{PRA_output} shows examples of the paths PRA generated for the relation \textit{biological-process-involves-gene-product} together with their weights. We only make use of relation paths with positive weights generated by PRA.

\begin{table}[bth!]
\small
\begin{center}
\begin{tabular}{|p{5.8cm}|l|}
\hline
\textbf{path} & \textbf{weight} \\ \hline
gene-encodes-gene-product(\textit{x,a}) $\wedge$ \_gene-plays-role-in-process(\textit{a,y}) & 10.53	\\\hline
\_isa(\textit{x,a}) $\wedge$ biological-process-involves-gene-product(\textit{a,y}) & 6.17 \\\hline
isa(\textit{x,a}) $\wedge$ biological-process-involves-gene-product(\textit{a,y}) & 2.80	\\\hline
gene-encodes-gene-product(\textit{x,a}) $\wedge$ \_gene-plays-role-in-process(\textit{a,b}) $\wedge$ isa(\textit{b,y}) & -0.06	\\\hline

\end{tabular}
\caption{Example PRA-induced paths and weights for the NCI relation \textit{biological-process-involves-gene-product}.}
\label{PRA_output}
\end{center}
\end{table}


The paths induced by PRA are used to identify potential false negatives in the negative training examples (Section \ref{sec:distantly_labelled_data}). Each negative training example is examined to check whether the entity pair is related in UMLS by following any of the relation paths extracted by PRA for the relevant target relation. Examples containing related entity pairs are assumed to be false negatives, since the relation can be inferred from the knowledge base, and removed from the set of negatives training examples. 
For instance, using the path in the top row of Table \ref{PRA_output}, sentences containing the entities $x$ and $y$ would be removed if the path \textit{gene-encodes-gene-product($x$,$a$) $\wedge$ \_gene-plays-role-in-process($a$,$y$)} could be identified within UMLS.




\subsection{Evaluation}

{\bf Relation Extraction system:} 
The MultiR system \cite{Hoffmann:2010} with features described by \newcite{Surdeanu:2011} was used for the experiments. \\
{\bf Datasets:} Three datasets were created to train MultiR and evaluate performance. The first ({\bf Unfiltered}) uses the data obtained using distant supervision (Section \ref{sec:distantly_labelled_data}) without removing any examples identified by PRA. The overall ratio of positive to negative sentences in this dataset was 1:5.1. However, this changes to 1:2.3 after removing examples identified by PRA. Consequently the bias in the distantly supervised data was adjusted to 1:2 to increase comparability across configurations. Reducing bias was also found to increase relation extraction performance, producing a strong baseline. 
The {\bf PRA-reduced} dataset is created by applying PRA reduction (Section \ref{sec:pra_reduction}) to the {\it Unfiltered} dataset to remove a portion of the negative training examples. Removing these examples produces a dataset that is smaller than {\it Unfiltered} and with a different bias.  Changing the bias of the training data can influence the classification results.  Consequently the {\bf Random-reduced} dataset was created by removing randomly selected negative examples from {\it Unfiltered} to produce a dataset with the same size and bias as {\it PRA-reduced}. 
The {\it Random-reduced} dataset is used to show that randomly removing negative instances leads to lower results than removing those suggested by PRA.
{\bf Evaluation:} Two approaches were used to evaluate performance.

The {\bf Held-out} datasets consist of the {\it Unfiltered}, {\it PRA-reduced} and {\it Random-reduced} data sets. 
The set of entity pairs obtained from the knowledge base is split into four parts and a process similar to 4-fold cross validation applied. In each fold the automatically labelled sentences obtained from the pairs in 3 of the quarters are used as training data and sentences obtained from the remaining quarter used for testing.

The {\bf Manually labelled} dataset contains 400 examples of the relation {\it may-prevent} and 400 of {\it may-treat} which were manually labelled by two annotators who were medical experts. Both relations are taken from the ND-FRT subset of UMLS. Each annotator was asked to label every sentence and then re-examine cases where there was disagreement. This process lead to inter-annotator agreement of 95.5\% for {\it may-treat} and 97.3\% for {\it may-prevent}. The annotated data set is publicly available\footnote{https://sites.google.com/site/umlscorpus/home}. Any sentences in the training data containing an entity pair that occurs within the manually labelled dataset are removed. Although this dataset is smaller than the held-out dataset, its annotations are more reliable and it is therefore likely to be a more accurate indicator of performance accuracy. 
This dataset is more balanced than the held-out data with a ratio of 1:1.3 for {\it may-treat} and 1:1.8 for {\it may-prevent}.


{\bf Evaluation metric:} Our experiments use entity level evaluation since this is the most appropriate approach to determine suitability for database population. Precision and recall are computed based on the proportion of entity pairs identified. For the held-out data the set of correct entity pairs are those which occur in sentences labeled as positive examples of the relation and which are also listed as being related in UMLS. For the manually labelled data it is simply the set of entity pairs that occur in positive examples of the relation. 


\section{Results}


\subsection{Held-out data}
	
Table \ref{held_out_results_MultiR} shows the results obtained using the held-out data. Overall results, averaged across all relations with maximum recall, are shown in the top portion of the table and indicate that applying PRA improves performance. Although the highest precision is obtained using the {\it Unfiltered} classifier, the {\it PRA-reduced} classifier leads to the best recall and F1. Performance of the {\it Random-reduced} classifier indicates that the improvement is not simply due to a change in the bias in the data but that the examples it contains lead to an improved model.



The lower part of Table \ref{held_out_results_MultiR} shows results for each relation. The \textit{PRA-reduced} classifier produces the best results for the majority of relations and always increases recall compared to {\it Unfiltered}.


\begin{table*}[bht!]
\small
\begin{center}
\begin{tabular}{|r|ccc|ccc|ccc|}
\hline

  & \multicolumn{3}{|c|}{\textbf{Unfiltered}} & \multicolumn{3}{|c|}{\textbf{Random-reduced}}  & \multicolumn{3}{|c|}{\textbf{PRA-reduced}}\\
\hline
& Prec. & Rec. & F1 & Prec. & Rec. & F1 & Prec. & Rec. & F1 \\
\hline
\hline
Overall & 62.30 & 51.82 & 56.58 & 44.49 & 74.26 & 55.64 & 56.85 & 77.10 & {\bf 65.44}\\
\hline
\hline
& \multicolumn{9}{|c|}{NCI relations} \\
\hline
biological\_process\_involves\_gene\_product  & 89.61 & 43.18 & 57.86 & 65.67 & 78.79 & 71.38 & 70.63 & 84.85 & \textbf{76.97} \\ 
disease\_has\_normal\_cell\_origin  & 60.20 & 83.86 & \textbf{69.95} & 43.2 & 95.21 & 58.85 & 42.80 & 91.88 & 57.91   \\ 
gene\_product\_has\_associated\_anatomy  & 41.65 & 64.04 & \textbf{49.96} & 29.22 & 74.63 & 41.81 & 37.94 & 65.28 & 47.82   \\ 
gene\_product\_has\_biochemical\_function  &  86.43 & 72.00 & 78.33 & 60.66 & 91.57 & 72.90 & 70.58 & 95.80 & \textbf{81.17}   \\ 
process\_involves\_gene  & 78.92 & 50.71 & 61.54 & 51.38 & 80.64 & 62.73 & 68.16 & 87.34 & \textbf{76.47}   \\ 
\hline
& \multicolumn{9}{|c|}{ND-FRT relations} \\
\hline
contraindicating\_class\_of  & 40.00 & 20.83 & 26.14 & 28.48 & 72.50 & 39.58 & 41.30 & 82.50 & \textbf{54.33}   \\ 
may\_prevent  & 27.48 & 14.69 & 18.87 & 20.61 & 44.79 & 27.94 & 38.11 & 35.63 & \textbf{36.64}   \\ 
may\_treat  & 48.66 & 39.63 & 43.14 & 39.57 & 50.00 & 43.84 & 50.88 & 57.93 & \textbf{54.11}   \\ 
mechanism\_of\_action\_of  & 47.15 & 40.63 & 43.12 & 40.25 & 59.38 & 47.62 & 52.85 & 59.38 & \textbf{55.82}   \\ 
\hline

\end{tabular}
\caption{Evaluation using held-out data} 
\label{held_out_results_MultiR}
\end{center}
\vspace{-0.1cm}
\end{table*}

\begin{table*}[t!]
\small
\begin{center}
\begin{tabular}{|r|ccc|ccc|ccc|}
\hline
 & \multicolumn{3}{|c|}{\textbf{Unfiltered}} & \multicolumn{3}{|c|}{\textbf{Random-reduced}}  & \multicolumn{3}{|c|}{\textbf{PRA-reduced}}\\
\hline
relation & Prec. & Rec. & F1 & Prec. & Rec. & F1 & Prec. & Rec. & F1  \\
\hline

may\_prevent & 54.17 & 21.67 & 30.95 & 53.57 & 25.00 & 34.09 & 39.66 & 38.33 & \textbf{38.98} \\ 
may\_treat & 40.00 & 47.48 & 43.42 & 43.21 & 50.36 & 46.51 & 41.05 & 67.63 & \textbf{51.09} \\ 
\hline
\end{tabular}
\caption{Evaluation using manually labelled data} 
\label{gold_standard_evaluation}
\end{center}
\vspace{-0.4cm}
\end{table*}

It is perhaps surprising that removing false negatives from the training data leads to an increase in recall, rather than precision. False negatives cause the classifier to generate an overly restrictive model of the relation and to predict positive examples of a relation as negative. Removing them leads to a less constrained model and higher recall. 

\begin{figure}[bht!] 
  \centering
     \scalebox{0.5}{\includegraphics[width=0.97\textwidth]{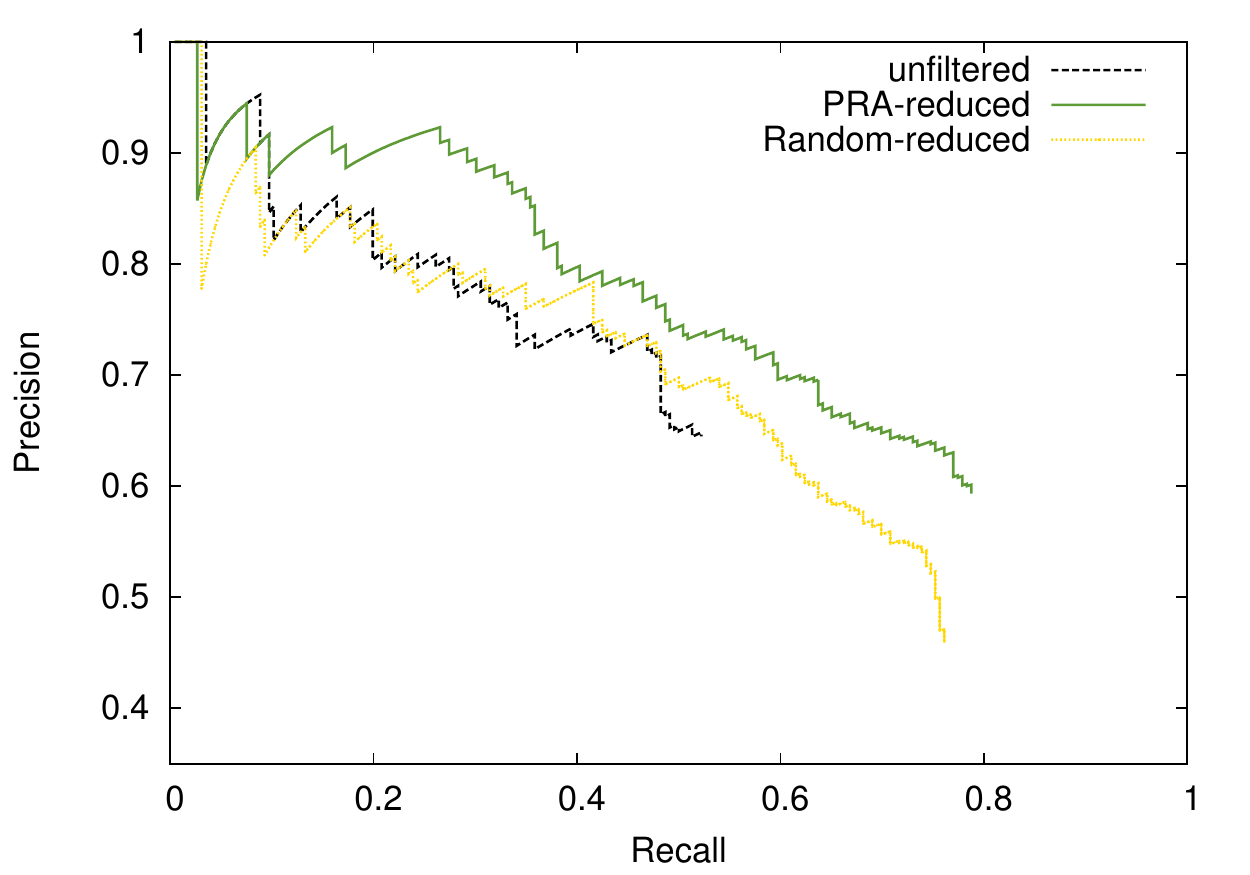}}
   \caption{Precision/Recall Curve for Held-out data}
  \label{fig:precision_recall_curve}
\end{figure}

There are two relations where there is also an increase in precision (\textit{contraindicating-class-of} and \textit{mechanism-of-action-of}) and these are also the ones for which the fewest training examples are available. The classifier has access to such a limited amount of data for these relations that removing the false negatives identified by PRA allows it to learn a more accurate model.


Figure \ref{fig:precision_recall_curve} presents a precision/recall curve computed using MultiR's output probabilities. Results for the \textit{PRA-reduced} and the \textit{Random-reduced} classifiers show that 
reducing the amount of negative training data increases recall. However, using \textit{PRA-reduced} generally leads to higher precision, indicating that PRA is able to identify suitable instances for removal from the training set. The {\it Unfiltered} classifier produces good results but precision and recall are lower than {\it PRA-reduced}.

\subsection{Manually labelled}

Table \ref{gold_standard_evaluation} shows results of evaluation on the more reliable manually labelled data set. The best overall performance is once again obtained using the {\it PRA-reduced} classifier. There is an increase in recall for both relations and a slight increase in precision for {\it may\_treat}. Performance of the {\it Random-reduced} classifier also improves due to an increasing recall but remains below {\it PRA-reduced}. Performance of the {\it Random-reduced} classifier is also better than {\it Unfiltered}, with the overall improvement largely resulting from increased recall, but below {\it PRA-reduced}. These results confirm that removing examples identified by PRA improves the quality of training data.

Further analysis indicated that the \textit{PRA-reduced} classifier produces the fewest false negatives in its predictions on the manually annotated dataset. It incorrectly labels 82 entity pairs (45 {\it may-treat}, 37 {\it may-prevent}) as negative while \textit{Unfiltered} predicts 120 (73, 47) and \textit{Random-reduced} 114 (69, 45). This supports our initial hypothesis that removing potential false negatives from training data improves classifier predictions. 



\section{Conclusions and Future Work}

This paper proposes a novel approach to identifying incorrectly labelled instances generated using distant supervision. Our method applies an inference learning method to detect and discard possible false negatives from the training data. We show that our method improves performance for a range of relations in the biomedical domain by making use of information from UMLS. 

In future we would like to explore alternative methods for selecting PRA relation paths to identify false negatives. Furthermore we would like to examine the PRA-reduced data in more detail. We would like to find which kind of entity pairs are detected by our proposed method and whether the reduced data can also be used to extend the positive training data. We would also like to apply the approach to other domains and alternative knowledge bases. Finally it would be interesting to compare our approach to other state of the art relation extraction systems for distant supervision or biased-SVM approaches such as \newcite{Liu:2003}.



\section*{Acknowledgements}

The authors are grateful to the Engineering and Physical Sciences Research Council for supporting the work described in this paper (EP/J008427/1).

\bibliographystyle{acl}
\bibliography{references}

\begin{thebibliography}{}

\bibitem[\protect\citename{Aronson and Lang}2010]{Aronson:2010}
A.~Aronson and F.~Lang.
\newblock 2010.
\newblock {An overview of MetaMap: historical perspective and recent advances}.
\newblock {\em Journal of the American Medical Association}, 17(3):229--236.

\bibitem[\protect\citename{Augenstein \bgroup et al.\egroup
  }2014]{Augenstein:2014b}
Isabelle Augenstein, Diana Maynard, and Fabio Ciravegna.
\newblock 2014.
\newblock Relation extraction from the web using distant supervision.
\newblock In {\em Proceedings of the 19th International Conference on Knowledge
  Engineering and Knowledge Management (EKAW 2014)}, Link\"{o}ping, Sweden,
  November.

\bibitem[\protect\citename{Bodenreider}2004]{Bod04}
Olivier Bodenreider.
\newblock 2004.
\newblock The unified medical language system (umls): integrating biomedical
  terminology.
\newblock {\em Nucleic acids research}, 32(suppl 1):D267--D270.

\bibitem[\protect\citename{Charniak and Johnson}2005]{Charniak:2005}
Eugene Charniak and Mark Johnson.
\newblock 2005.
\newblock Coarse-to-fine n-best parsing and maxent discriminative reranking.
\newblock In {\em Proceedings of the 43rd Annual Meeting on Association for
  Computational Linguistics}, ACL '05, pages 173--180, Stroudsburg, PA, USA.
  Association for Computational Linguistics.

\bibitem[\protect\citename{Cohen and Hunter}2013]{Coh13}
K~Bretonnel Cohen and Lawrence~E Hunter.
\newblock 2013.
\newblock Text mining for translational bioinformatics.
\newblock {\em PLoS computational biology}, 9(4):e1003044.

\bibitem[\protect\citename{Craven and Kumlien}1999]{Craven:1999}
Mark Craven and Johan Kumlien.
\newblock 1999.
\newblock Constructing biological knowledge bases by extracting information
  from text sources.
\newblock In {\em In Proceedings of the Seventh International Conference on
  Intelligent Systems for Molecular Biology (ISMB)}, pages 77--86. AAAI Press.

\bibitem[\protect\citename{Hahn \bgroup et al.\egroup }2012]{Hah12}
Udo Hahn, K~Bretonnel Cohen, Yael Garten, and Nigam~H Shah.
\newblock 2012.
\newblock Mining the pharmacogenomics literature—a survey of the state of the
  art.
\newblock {\em Briefings in bioinformatics}, 13(4):460--494.

\bibitem[\protect\citename{Hoffmann \bgroup et al.\egroup }2010]{Hoffmann:2010}
Raphael Hoffmann, Congle Zhang, and Daniel~S. Weld.
\newblock 2010.
\newblock Learning 5000 relational extractors.
\newblock In {\em Proceedings of the 48th Annual Meeting of the Association for
  Computational Linguistics}, ACL '10, pages 286--295, Stroudsburg, PA, USA.
  Association for Computational Linguistics.

\bibitem[\protect\citename{Intxaurrondo \bgroup et al.\egroup
  }2013]{Intxaurrondo:2013}
Ander Intxaurrondo, Mihai Surdeanu, Oier~Lopez de~Lacalle, and Eneko Agirre.
\newblock 2013.
\newblock Removing noisy mentions for distant supervision.
\newblock {\em Procesamiento del Lenguaje Natural}, 51:41--48.

\bibitem[\protect\citename{Jensen \bgroup et al.\egroup }2006]{Jen06}
Lars~Juhl Jensen, Jasmin Saric, and Peer Bork.
\newblock 2006.
\newblock Literature mining for the biologist: from information retrieval to
  biological discovery.
\newblock {\em Nature reviews genetics}, 7(2):119--129.

\bibitem[\protect\citename{Krause \bgroup et al.\egroup }2012]{Krause:2012}
Sebastian Krause, Hong Li, Hans Uszkoreit, and Feiyu Xu.
\newblock 2012.
\newblock Large-scale learning of relation-extraction rules with distant
  supervision from the web.
\newblock In {\em Proceedings of the 11th International Conference on The
  Semantic Web - Volume Part I}, ISWC'12, pages 263--278, Berlin, Heidelberg.
  Springer-Verlag.

\bibitem[\protect\citename{Lao and Cohen}2010]{Lao:2010}
Ni~Lao and William~W. Cohen.
\newblock 2010.
\newblock Relational retrieval using a combination of path-constrained random
  walks.
\newblock {\em Mach. Learn.}, 81(1):53--67, October.

\bibitem[\protect\citename{Lao \bgroup et al.\egroup }2011]{Lao:2011}
Ni~Lao, Tom Mitchell, and William~W. Cohen.
\newblock 2011.
\newblock Random walk inference and learning in a large scale knowledge base.
\newblock In {\em Proceedings of the 2011 Conference on Empirical Methods in
  Natural Language Processing}, pages 529--539, Edinburgh, Scotland, UK., July.
  Association for Computational Linguistics.

\bibitem[\protect\citename{Liu \bgroup et al.\egroup }2003]{Liu:2003}
Bing Liu, Yang Dai, Xiaoli Li, Wee~Sun Lee, and Philip~S. Yu.
\newblock 2003.
\newblock Building text classifiers using positive and unlabeled examples.
\newblock In {\em Intl. Conf. on Data Mining}, pages 179--188.

\bibitem[\protect\citename{Min \bgroup et al.\egroup }2013]{Min13}
Bonan Min, Ralph Grishman, Li~Wan, Chang Wang, and David Gondek.
\newblock 2013.
\newblock Distant supervision for relation extraction with an incomplete
  knowledge base.
\newblock In {\em Proceedings of the 2013 Conference of the North American
  Chapter of the Association for Computational Linguistics: Human Language
  Technologies}, pages 777--782, Atlanta, Georgia, June. Association for
  Computational Linguistics.

\bibitem[\protect\citename{Mintz \bgroup et al.\egroup }2009]{Mintz:2009}
Mike Mintz, Steven Bills, Rion Snow, and Dan Jurafsky.
\newblock 2009.
\newblock Distant supervision for relation extraction without labeled data.
\newblock In {\em Proceedings of the Joint Conference of the 47th Annual
  Meeting of the ACL and the 4th International Joint Conference on Natural
  Language Processing of the AFNLP: Volume 2 - Volume 2}, ACL '09, pages
  1003--1011, Stroudsburg, PA, USA. Association for Computational Linguistics.

\bibitem[\protect\citename{Nguyen and Moschitti}2011]{Nguyen:2011a}
Truc-Vien~T. Nguyen and Alessandro Moschitti.
\newblock 2011.
\newblock End-to-end relation extraction using distant supervision from
  external semantic repositories.
\newblock In {\em Proceedings of the 49th Annual Meeting of the Association for
  Computational Linguistics: Human Language Technologies: short papers - Volume
  2}, HLT '11, pages 277--282, Stroudsburg, PA, USA. Association for
  Computational Linguistics.

\bibitem[\protect\citename{Porter}1997]{Porter:1997}
M.~F. Porter.
\newblock 1997.
\newblock Readings in information retrieval.
\newblock chapter An Algorithm for Suffix Stripping, pages 313--316. Morgan
  Kaufmann Publishers Inc., San Francisco, CA, USA.

\bibitem[\protect\citename{Riedel \bgroup et al.\egroup }2010]{Riedel:2010}
Sebastian Riedel, Limin Yao, and Andrew McCallum.
\newblock 2010.
\newblock Modeling relations and their mentions without labeled text.
\newblock In {\em Proceedings of the European Conference on Machine Learning
  and Knowledge Discovery in Databases (ECML PKDD '10)}.

\bibitem[\protect\citename{Ritter \bgroup et al.\egroup }2013]{Rit13}
Alan Ritter, Luke Zettlemoyer, Oren Etzioni, et~al.
\newblock 2013.
\newblock Modeling missing data in distant supervision for information
  extraction.
\newblock {\em Transactions of the Association for Computational Linguistics},
  1:367--378.

\bibitem[\protect\citename{Roller and Stevenson}2014]{Roller:2014b}
Roland Roller and Mark Stevenson.
\newblock 2014.
\newblock Self-supervised relation extraction using umls.
\newblock In {\em Proceedings of the Conference and Labs of the Evaluation
  Forum 2014}, Sheffield, England.

\bibitem[\protect\citename{Surdeanu \bgroup et al.\egroup }2011]{Surdeanu:2011}
Mihai Surdeanu, David McClosky, Mason Smith, Andrey Gusev, and Christopher
  Manning.
\newblock 2011.
\newblock Customizing an information extraction system to a new domain.
\newblock In {\em Proceedings of the ACL 2011 Workshop on Relational Models of
  Semantics}, pages 2--10, Portland, Oregon, USA, June. Association for
  Computational Linguistics.

\bibitem[\protect\citename{Surdeanu \bgroup et al.\egroup }2012]{Surdeanu:2012}
Mihai Surdeanu, Julie Tibshirani, Ramesh Nallapati, and Christopher~D. Manning.
\newblock 2012.
\newblock Multi-instance multi-label learning for relation extraction.
\newblock In {\em Proceedings of the 2012 Joint Conference on Empirical Methods
  in Natural Language Processing and Computational Natural Language Learning},
  EMNLP-CoNLL '12, pages 455--465, Stroudsburg, PA, USA. Association for
  Computational Linguistics.

\bibitem[\protect\citename{Takamatsu \bgroup et al.\egroup
  }2012]{Takamatsu:2012}
Shingo Takamatsu, Issei Sato, and Hiroshi Nakagawa.
\newblock 2012.
\newblock Reducing wrong labels in distant supervision for relation extraction.
\newblock In {\em Proceedings of the 50th Annual Meeting of the Association for
  Computational Linguistics: Long Papers - Volume 1}, ACL '12, pages 721--729,
  Stroudsburg, PA, USA. Association for Computational Linguistics.

\bibitem[\protect\citename{Zhang \bgroup et al.\egroup }2013]{Zhang:2013}
Xingxing Zhang, Jianwen Zhang, Junyu Zeng, Jun Yan, Zheng Chen, and Zhifang
  Sui.
\newblock 2013.
\newblock Towards accurate distant supervision for relational facts extraction.
\newblock In {\em Proceedings of the 51st Annual Meeting of the Association for
  Computational Linguistics (Volume 2: Short Papers)}, pages 810--815, Sofia,
  Bulgaria, August. Association for Computational Linguistics.

\end{thebibliography}

\end{document}